\documentclass[runningheads]{llncs}
\usepackage{graphicx}
\usepackage{amsmath,amssymb} 
\usepackage{color}

\begin{document}
\title{An empirical study towards understanding how deep convolutional nets recognize falls} 

\titlerunning{empirical study for deep fall recognition}
%
\author{Yan Zhang\inst{1} \and
Heiko Neumann\inst{1}}
%
\authorrunning{Y. Zhang and H. Neumann}
%

\institute{Institute of Neural Information Processing, Ulm University, Germany \\
\email{\{yan.zhang,heiko.neumann\}@uni-ulm.de}}
\maketitle              
\begin{abstract}
Detecting unintended falls is essential for ambient intelligence and healthcare of elderly people living alone. In recent years, deep convolutional nets are widely used in human action analysis, based on which a number of fall detection methods have been proposed. Despite their highly effective performances, the behaviors of how the convolutional nets recognize falls are still not clear. In this paper, instead of proposing a novel approach, we perform a systematical empirical study, attempting to investigate the underlying fall recognition process. We propose four tasks to investigate, which involve five types of input modalities, seven net instances and different training samples. The obtained quantitative and qualitative results reveal the patterns that the nets tend to learn, and several factors that can heavily influence the performances on fall recognition. We expect that our conclusions are favorable to proposing better deep learning solutions to fall detection systems.

\end{abstract}

\section{Introduction}
Due to cognitive impairment or deficiencies of motor functionalities, unintended falls occur frequently in the group of elderly people, and can lead to severe or even fatal injuries \cite{dykes2010fall} \cite{gillain2014falls}. Therefore, to build up fall detection systems for elderly people healthcare, it is essential to recognize falls in an automatic and effective manner.

Fall recognition has been intensively studied in the past. If the human body dynamics has been precisely measured, identifying an unintended fall is straightforward. For example, one can recognize falls via measuring the vertical velocity of the human body towards the ground. If the velocity is above a threshold, then a fall occurs. Consequently, researchers tend to propose novel solutions to capture the body configurations and motions. For example, the work of \cite{zhang2006fall} uses a wearable triaxial accelerometer to measure the body motion and recognizes falls via one-class support vector machine. The work of \cite{wu2015development} develops a wearable system (mainly based on the accelerometer and GPS) to detect and localize falls in the wild. Wearable sensors enable measuring physical attributes of the human body in a precise and real-time manner. However, the sensors have to be physically attached to people, causing obstructive interventions to their daily living activities.

Computer vision technologies realize non-obstructive measurement of human body motions and conduct behavior recognition only based on imagery data. The effectiveness is highly improved when deep convolutional networks trained on large-scale image datasets are employed. To recognize a fall, two families of methods can be considered: The first family attempts to capture precise body configurations over time, such as \cite{cao2017realtime} and \cite{insafutdinov2017cvpr} for 2D pose estimation and \cite{guler2018densepose} for 3D pose estimation. Such pose estimation methods can replace the functionality of wearable sensors but perform human body measurement in a non-contacting manner. The second family, which is usually based on deep convolutional nets, aims at inferring the semantic content of the input data via creating a mapping from the input data to the action labels in an end-to-end fashion. For example, \cite{simonyan2014two} yields an action label for an input sequence, \cite{neverova2014multi} yields both action labels and temporal durations, and \cite{Lea_2017_CVPR} produces frame-wise labels for temporal action segmentation. In this paper, we focus on the second family of methods, since the end-to-end inference behavior does not need any intermediate step, e.g., training a classifier based on the captured body poses. In addition, the data annotation procedure only requires to assign action labels to frames/videos, instead of annotating the key joints on the human bodies in each frame as the first family of methods.

Although several relevant methods like \cite{nunez2017vision} have been proposed, the underlying reasons of the effectiveness are still not clear. In this paper, rather than proposing a novel method for fall recognition, we aim at attaining insights of how the deep convolutional net recognizes falls via a series of empirical investigations. Our study is based on a family of convolutional encoder-decoder nets, different types of input modalities and recordings from different environments. According to our investigations, we discover: (1) Human body motion represented by the optical flow is highly informative for the net to recognize falls. (2) The net tends to learn human body-centered context, namely the appearance surrounding the human body, if the training samples have RGB frames. However, the net cannot get rid of the influence of the background context irrelevant to falling, and lacks generalizability across different environments. (3) The human-centered context and human body motion are complementary. (4) Inaccurate body pose information can degrade the performances.

{\bf Organization.} This paper is organized as follows. Section \ref{sec:related_work} introduces related work on vision-based methods for fall recognition and work on model explanation. Section \ref{sec:methods} introduces the employed convolutional net, as well as different sorts of attribute maps for model explanation. In Section \ref{sec:experiments}, we present our empirical investigations, results and discussions. In the end, we conclude our work and propose future studies in Section \ref{sec:conclusion}.

\section{Related Work}
\label{sec:related_work}
Systematic reviews of fall recognition and detection systems can be found in \cite{igual2013challenges} and \cite{mubashir2013survey}, which cover solutions based on diverse types of sensors. For vision-based methods, a typical processing pipeline consists of background subtraction, feature extraction and classification, as presented in \cite{rougier2011robust} \cite{vishwakarma2007automatic} and others. Each step in this pipeline is normally hand-crafted, separately considered and implemented based on certain heuristic rules. A frequently considered rule is that the background information is redundant for fall detection. Thus, background subtraction is performed by algorithms like training Gaussian mixture models, subspace clustering or other sophisticated approaches \cite{piccardi2004background}. Another heuristic rule is that the body shape is a pronounced feature of falling. Consequently, the silhouette of the human body \cite{rougier2011robust} \cite{anderson2006recognizing}, or the shape of the foreground bounding box \cite{vishwakarma2007automatic} \cite{toreyin2005hmm}, is extracted and analyzed. Nevertheless, heuristics are not always precise and comprehensive. The studies of \cite{stone2015fall} and \cite{solbach2017vision} present effective fall detection solutions when considering the ground plane, indicating that the background information can be very useful.

Comparing with traditional vision-based approaches, deep learning methods enable end-to-end inference with minimal pre-processing on the input data, and the deep nets can learn representative features from the data automatically. Therefore, the algorithm is not necessary to rely on non-guaranteed heuristics. Several studies report that deep learning methods lead to better performances in terms of action recognition \cite{simonyan2014two} \cite{Carreira_2017_CVPR}, action detection \cite{yeung2016end} \cite{gkioxari2015finding} \cite{shou2017cdc}, action parsing \cite{Lea_2017_CVPR} \cite{lea2016temporal} and other tasks of human behavior analysis. Their success encourages many studies of fall recognition based on deep neural networks. For example, the work of \cite{nunez2017vision} employs a convolutional net with a similar architecture to the VGG-16 net \cite{simonyan2014very} and uses optical flow as the input modality. The work of \cite{wang2016human} uses a PCANet to recognize falls from image sequences with the assistance of foreground detection.

To understand the behaviors of deep convolutional nets, several types of attribute maps have been proposed \cite{babiker2017introduction} \cite{ancona2017unified}. For a specific input and a target class, the attribute map has the same spatial resolution with the input, and reveals the influence of each input pixel to the probability of the target class. The work of \cite{simonyan2013deep} proposes a saliency map, which is computed as the derivative of the output with respect to the input. \cite{sundararajan2017axiomatic} proposes the {integrated gradients}, in which the values show the difference between the net output of a reference input (normally zero) and the net output of a sample. \cite{shrikumar2017learning} proposes the DeepLIFT attribute measure, which can be regarded as an approximated version of {integrated gradients} according to \cite{ancona2017unified}.

\section{The convolutional net}
\label{sec:methods}

\begin{figure}
\centering
\includegraphics[height=6.5cm]{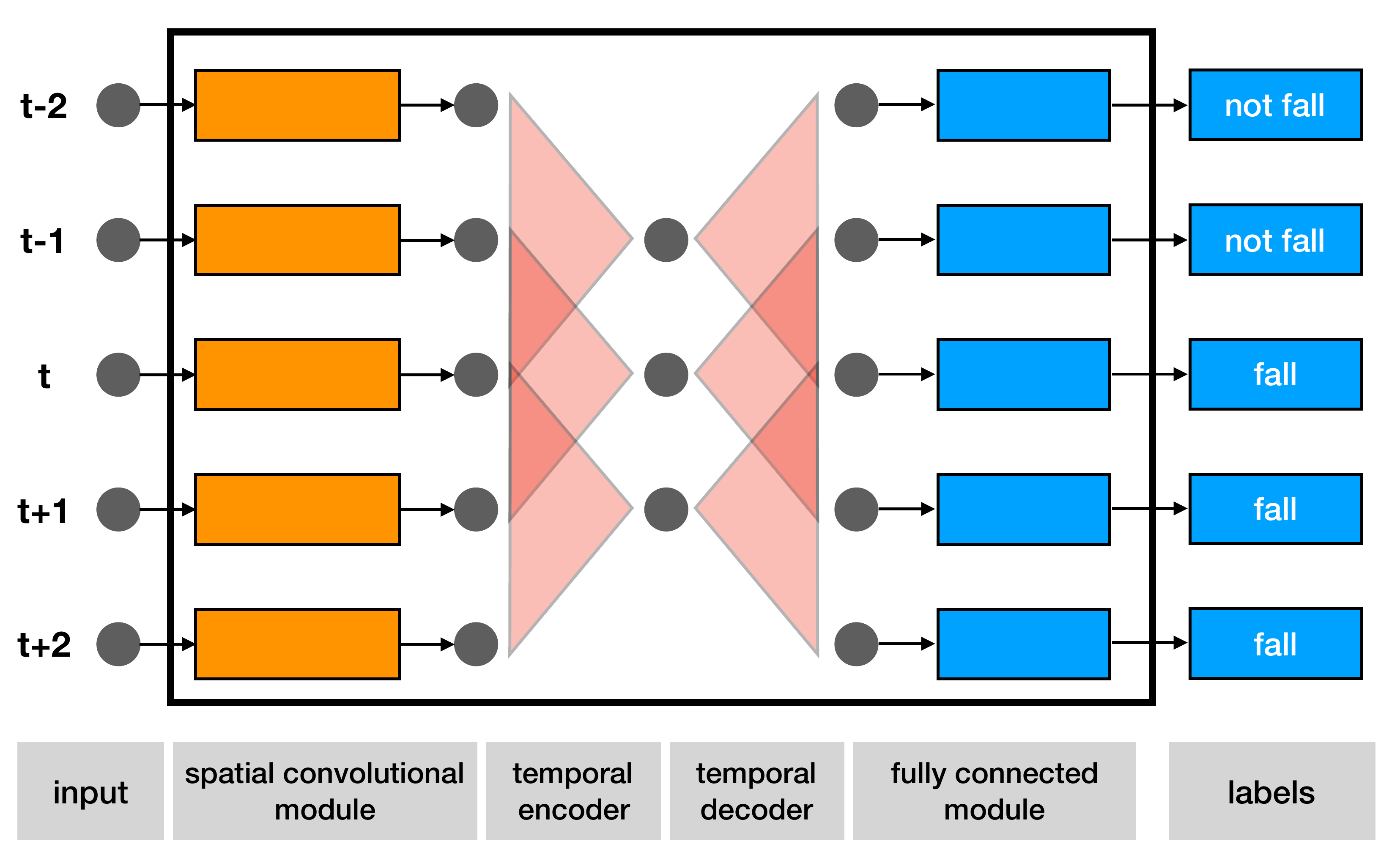}
\caption{The architecture of the convolutional encoder-decoder (CED) net. In the input layer, each frame (the gray node) is a 3D tensor with [height, width, channels].}
\label{fig:architecture}
\end{figure}

\begin{figure}
\centering
\includegraphics[height=8cm]{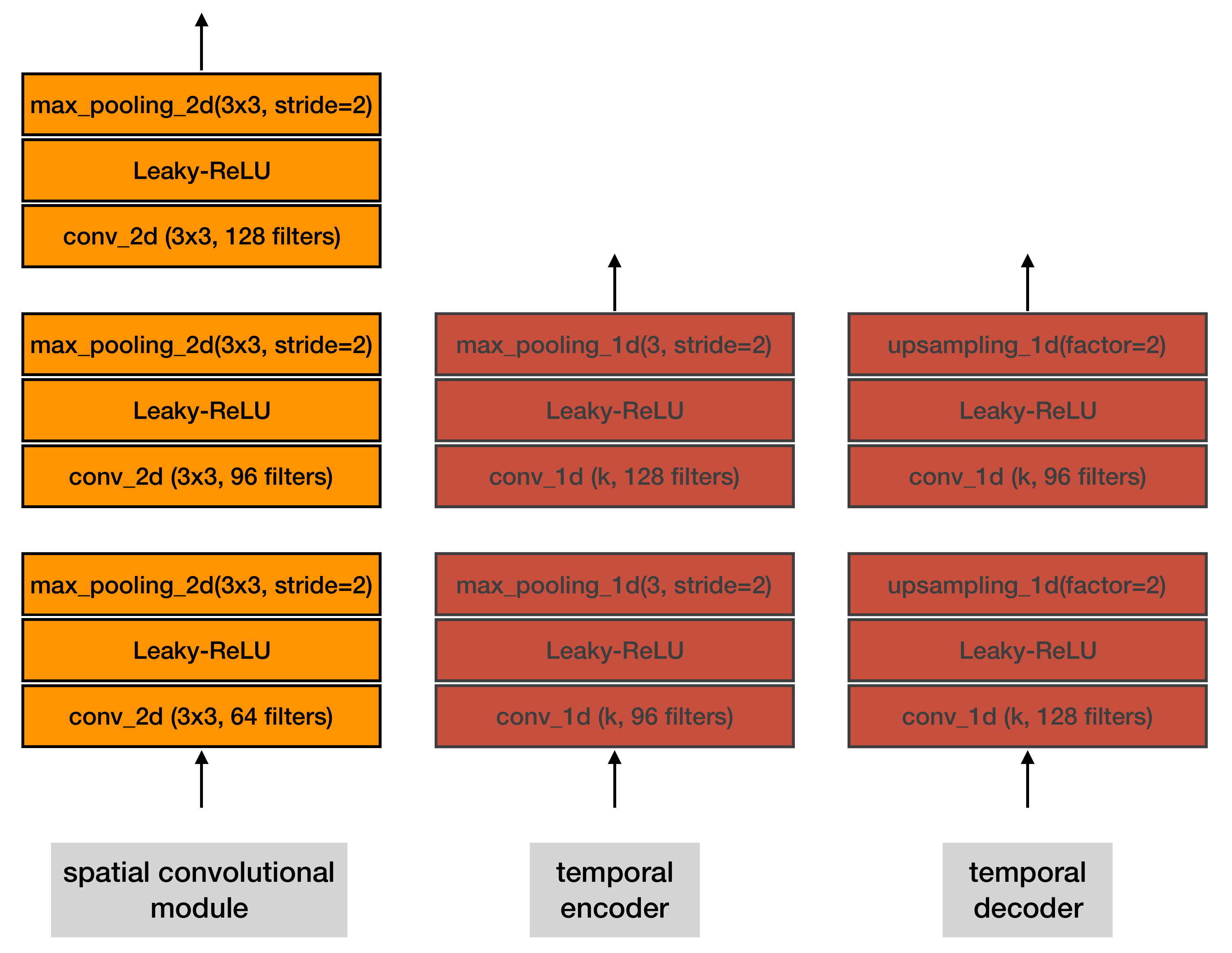}
\caption{Module specifications of our CED net, in which the data flows from the bottom to the top. The value of $k$ is determined in Section \ref{sec:experiments}. }
\label{fig:net_specification}
\end{figure}

We formulate fall recognition as a binary classification problem, and expect to obtain frame-wise semantic labels, so that recognition and temporal localization can be solved simultaneously. Therefore, we use a convolutional encoder-decoder (CED) architecture, which is modified from the non-causal ED-TCN model \cite{Lea_2017_CVPR}. Comparing with \cite{Lea_2017_CVPR}, our CED net combines the spatial net and the temporal net into a coherent structure. The architecture is illustrated in Figure \ref{fig:architecture} and the specifications are presented in Figure \ref{fig:net_specification}.

The CED architecture has several advantages besides generating frame-wise labels: (1) The CED net can capture long-range temporal dependencies, and outperforms recurrent nets, e.g. bidirectional LSTMs \cite{graves2005bidirectional} \cite{singh2016multi}, w.r.t. temporal action segmentation \cite{Lea_2017_CVPR} and motion prediction \cite{li2018convolutional}. (2) The CED net can generate piece-wise constant label sequences directly, without post-processing steps like median filtering. (3) Comparing with recurrent neural nets, in our trials we find that CED is much easier to train and converges much faster. (4) Once CED is trained, the model can process sequences of arbitrary lengths. Because of such merits, we only consider the convolutonal net in our study, and investigating recurrent neural nets is beyond our scope.

The CED model consists of several modules as shown in Figure \ref{fig:architecture}. In the following content, we introduce each of them.

\subsection{The spatial convolutional module}
Our convolutional module aims at extracting the feature of each individual frame in the video. It consists of three convolutional blocks, and each block contains a 2D convolutional layer, an activation function layer and a 2D max-pooling layer, following the architecture of the VGG net \cite{simonyan2014very} \cite{Lea_2017_CVPR}. After each block, the spatial resolution is downsampled by the factor of 2. At the end of the module, the input 3D tensor is flattened to a 1D vector. The specifications of the spatial convolutional module are shown in Figure \ref{fig:net_specification}. The number of convolution filters are suggested by \cite{lea2016segmental}. In our work, we use the leaky-ReLU \cite{maas2013rectifier} activation function, due to the superior performances to standard to the ReLU function, as indicated in \cite{xu2015empirical}. Moreover, the spatial convolution module is applied on each individual frame of the input tensor sequence, and has shared parameters across all frames.

\subsection{The temporal encoder and decoder}
After the spatial convolutional module, the 3D tensor of each frame converts to a 1D vector, and then all the vectors are stacked along the temporal dimension to compose a 2D tensor with the shape of [time, dimension] (or a 3D tensor with the shape of [batch, time, dimension]).

Similar to the 2D convolution operation, which can effectively capture spatial local features, the 1D temporal convolution computes temporal correlations between frames, in which the value of the kernel size $k$ specifies the size of the receptive field. The 1D max pooling operation downsamples the data along the time dimension to yield a compressed data representation. On the other and, the upsampling operation increases the temporal resolution to recover the original time length. The encoder and decoder have symmetric architectures, and hence the temporal encoder input and the temporal decoder output has the same temporal length.

\subsection{The fully connected module}
The fully connected module consists of a fully connected layer, a dropout layer and a softmax layer, and is applied on individual frames in the output of temporal decoder with shared parameters. Due to our binary classification setting, the output dimension of the fully connected layer can be 1 or 2. Here we use the two-dimensional output, since we expect that the insights derived from our work can be extended to multi-class classification problems straightforwardly. The dropout layer (with a keep ratio of 0.5) is used to avoid overfitting, and the softmax layer converts the scores to probabilities.

\subsection{Training the network}
In our work, all the modules are trained jointly, in contrast to \cite{Lea_2017_CVPR} that only trains the temporal encoder-decoder using the outputs from a pre-trained spatial net. For each frame, we compute the {\em cross-entropy} between the one-hot encoded ground truth label and the softmax output. Then the loss of the sequence is the sum of the cross-entropy values of all frames. After specifying the loss, the model parameters are learned via the Adam algorithm \cite{kingma2014adam}. Comparing with the stochastic gradient descent method, Adam can lead to superior results as reported in \cite{kingma2014adam}. In addition, the adaptive momentum nature is suitable for our problems, since our input modality can cause sparse gradients, like optical flows with motion information only on the foreground. Implementation details refer to Section \ref{sec:experiments}.

\section{Experiments}
\label{sec:experiments}
In this section, we present our empirical experiments to investigate how the deep convolutional net CED recognize falls. We propose 4 tasks, and for each task the quantitative results are shown by cross-validated frame-wise accuracies and the qualitative results are shown by attribute maps.

\subsection{Dataset}
\label{subsec:dataset}
We use the {\bf Le2i} Fall detection dataset presented in \cite{charfi2013optimized}, which is built using a single camera in realistic surveillance setting containing illumination variations, occlusions by furnitures, different appearances of the subjects, different types of falls (e.g. falling forward, falling while sitting, etc.), and other factors that simulate falls in daily lives. The video has spatial resolution of 320$\times$240 of pixels and is captured with 25 fps. Each video is annotated in a frame-wise fashion, which fits the CED architecture.  

The original dataset contains 4 environments, i.e. {\em home}, {\em lecture room}, {\em coffee room} and {\em office}. Due to loss of annotation files, we only use the recordings from {\em home} and {\em coffee room} in our study. For each of the two environments, there exist two groups of recordings.

\begin{figure}
\centering
\includegraphics[height=3.5cm]{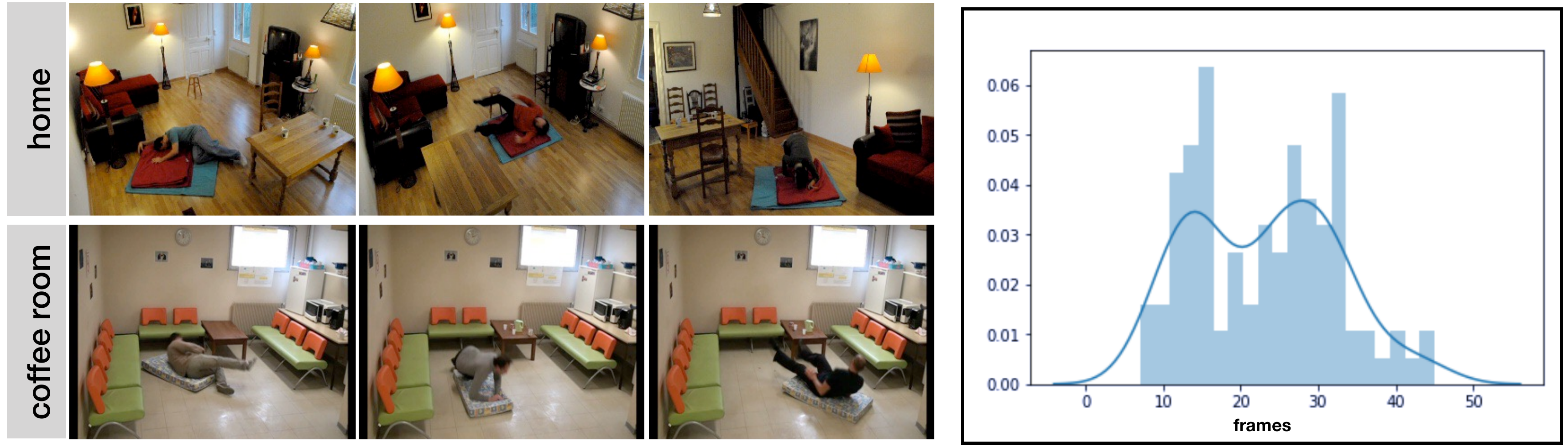}
\caption{From left to right: (1) Sample frames of falls in {\em home} and {\em coffee room}. (2) The statistics of time durations of falls across all videos, in which the x-axis denotes the fall duration, the y-axis and the bins show the normalized occurrence frequencies and the curve shows the fitted distribution.}
\label{fig:dataset_samples}
\end{figure}

\paragraph{Data preparation.} Since we focus on frame-wise fall recognition, in order to balance the number of fall and not-fall frames, from each video containing falling we extract a video snippet consisting of frames before, during and after the fall. Video trimming is based on the statistics of time durations of falls, which is shown in Figure \ref{fig:dataset_samples}. Specifically, the extracted snippet has 60 frames (2.4 seconds), starting from $T-49$ to $T+10$, where $T$ is the timestamp of the last frame of fall in the video. 

Depending on the recording environment, we perform a {\em high-level} splitting to divide the dataset into 2 folds, each of which contains recordings from either {\em home} or {\em coffee room}. Since there are two groups for each environment, we perform a {\em low-level} splitting to divide the dataset into 4 folds. Therefore, the {\em high-level} splitting can be used for cross-environment validation, and the {\em low-level} splitting can be used for cross-validation under small environment variations. 

After such preparation step, we obtain a new dataset incorporating 99 video snippets with 2 {\em high-level} splits and 4 {\em mid-level} splits.

\subsection{Input modalities to the net}
Besides the RGB frames, we also compute time differences, TV-L1 optical flows \cite{chambolle2004algorithm}, and score maps of human body poses\footnote{The MPII body model has 14 keypoints and hence the method generates 14 pose score maps for each image. In our experiment, we average these 14 score maps to one map.} \cite{insafutdinov2016eccv} \cite{insafutdinov2017cvpr} as the net input modalities. For computational purposes, we downsample the spatial resolution to $56 \times 56$.

Similar to \cite{Lea_2017_CVPR}, each frame of the net input contains a stack of frames from the original data sequence. Denoting the {\em standardized} RGB image sequence as $\{I_t\}$, the optical flow sequence as $\{w_t\}$ (values within $[-20,20]$) and the sequence of score maps as $\{s_t\}$ (values within $[0,1]$), the modalities used in our experiments are shown in Table \ref{tab:modalities}. 

\begin{table}
\centering
\caption{The input modalities used in our experiments, in which the {\bf Pose+Optical Flow} modality uses the normalized optical flow $\tilde{w}_{t}$.}
\label{tab:modalities}
\begin{tabular}{l|l}

{\sf Modalities} & {\sf Format of each frame}\\
\hline
{\bf RGB+TimeDifference} & $\{ I_{t-1},I_{t},I_{t+1}, I_{t}-I_{t-1}, I_{t+1}-I_{t} \}$\\
{\bf TimeDifference} &  $\{ I_{t}-I_{t-1}, I_{t+1}-I_{t} \}$\\
{\bf Optical Flow} &  $\{  w_{t-1}, w_{t}, w_{t+1} \}$\\
{\bf Pose} & $\{ s_{t-1}, s_{t}, s_{t+1} \}$\\
{\bf Pose+Optical Flow} & $\{ s_{t-1}, \tilde{w}_{t-1}, s_t, \tilde{w}_t, s_{t+1}, \tilde{w}_{t+1} \}$\\
\end{tabular}

\end{table}

The {\bf RGB+TimeDifference} modality is suggested by \cite{Lea_2017_CVPR}, in which the RGB frames encode the appearances of the visual scene and the time differences have the functionality of attention. Image standardization is performed frame-wisely, in order to eliminate the influence of illumination changes. Since the background is static, {\bf TimeDifference} and {\bf Optical Flow} focus on the human body, while {\bf TimeDifference} does not incorporate directional human body motions. The pose information is represented by the score map produced by the pre-trained model of \cite{insafutdinov2016eccv} \cite{insafutdinov2017cvpr}, which is beneficial for person re-identification and tracking \cite{tang2017multiple}. When combining optical flow and pose, we expect that the pose score map works as an attention mechanism, encouraging the net to learn motion features around the body key points. 

One can note that the {\bf Pose+Optical Flow} modality uses the normalized optical flow $\tilde{w}_{t}$, which is computed by $w_t/20$ and hence ranges within $[-1,1]$. In this case, the ranges of the pose score map and the optical flow are similar. We find that such flow normalization process is beneficial in our trials. A probable reason is that the normalization leads to similar ranges of convolution parameters for the flow and the pose map in {\bf Pose+Optical Flow}.

\subsection{Implementation}
The implementation is based on Tensorflow. The batch size is fixed to 8, meaning 8 tensor sequences are fed to the net for one iteration. The Adam algorithm is used to train the model \cite{kingma2014adam}, where the initial learning rate is $0.001$ and other parameters are set to the Tensorflow default values. The learning rate is decayed every 10 epochs, namely $0.001\times 0.9^{\lfloor \frac{epoch}{10}\rfloor}$, and training terminates after 100 epochs. In our trials, more iterations lead to comparable or worse results. 

In addition, attribute map extraction is implemented based on the DeepExplain library introduced in \cite{ancona2017unified}. 

\subsection{Evaluation methods}
Rather than performing model selection as in \cite{Lea_2017_CVPR}, we use a family of net instances to verify whether some conditions can consistently influence the performances. We vary two influential factors in the net architecture, i.e., the temporal convolution kernel size $k$ determining the temporal receptive field, and the temporal length of the input sequence $l$ determining the up-limit range of the temporal structure that the net can learn. In our experiments, we use the net instances with $(k,l)\in \{(3,8), (3,16), (3,32), (5,16), (5,32), (7,16), (7,32)\}$.

For the {\em high-level} splitting, 2-fold cross-validation is performed, in which each net instance is trained on the first fold and validated on the second, and vice versa. Then, for each net instance, the two validated accuracies are averaged to derive the cross-validated accuracy. For the {\em low-level} splitting, 4-fold cross-validation is performed in an identical manner. Since each net instance associates with an accuracy value, the quantitative performance of the CED model is presented in terms of a box plot. 

The qualitative results are shown by attribute maps, i.e. DeepLIFT \cite{shrikumar2017learning}, integrated gradients \cite{ancona2017unified} and saliency maps \cite{simonyan2013deep}. In addition, each attribute map is stacked to the map of edges of the input for visualization purposes.

\subsection{Tasks, results and discussions}

\subsubsection{Task 1: Investigating the cross-environment generalizability}
\begin{figure}
\centering
\includegraphics[height=6.2cm]{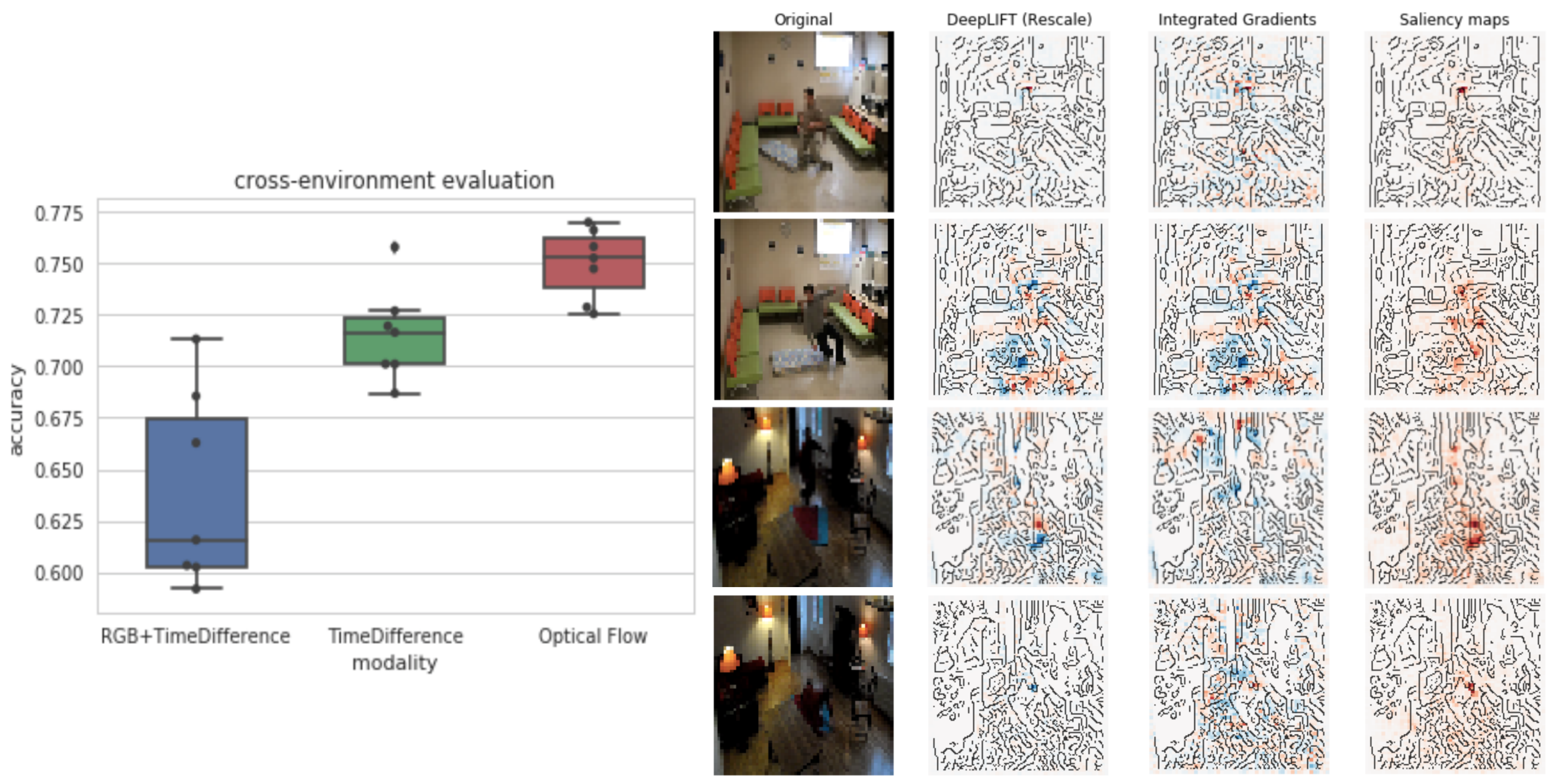}
\caption{From left to right: (1) The quantitative results of the 2-fold cross-validation, where the results from each net instance are shown as black dots in parallel to the box plots. In each box plot, the bar inside the box denotes the median, and the box shows the interquartile range (IQR) and the samples between whiskers with $1.5 \times$IQR are inliers. (2) The attribute maps of frames from four testing recordings are shown, where the red color and the blue color denote contribution and suppression effects on the probability of falling.}
\label{fig:cross_environment}
\end{figure}

In this task, we aim at investigating the generalizability across environments, namely, how the CED performs if training samples and testing samples are collected from totally different environments. Therefore, we conduct a 2-fold cross-validation procedure based on the {\em high-level} splitting, and use {\bf RGB+TimeDifference}, {\bf TimeDifference} and {\bf Optical Flow} as the input modalities. The results are shown in Figure \ref{fig:cross_environment}.

From the box plots, one can see that {\bf RGB+TimeDifference} performs inferior to {\bf TimeDifference} and {\bf Optical Flow}, and {\bf Optical Flow} outperforms {\bf TimeDifference}. In addition, the attribute maps from four testing recordings consistently show that many pixels on the background can heavily affect the net inference process.

\paragraph{Discussion}
The net with {\bf RGB+TimeDifference} performs just slightly better than random guess, due to the binary classification setting. The attribute maps show that irrelevant background information has strong influence on fall recognition, and hence we consider that the net cannot discard irrelevant background information automatically during training, and leads to degraded generalizability across environments. Excluding the background information, as in {\bf TimeDifference} and {\bf Optical Flow}, can improve the performances dramatically. This fact can indicate that real influential and environment-invariant features of falls are human body-centered. In addition, the superior performances of {\bf Optical Flow} to {\bf TimeDifference} can indicate that the directional body motion contains more representative information of falls.

\subsubsection{Task 2: Investigating the influence of training samples}
\begin{figure}
\centering
\includegraphics[height=6.2cm]{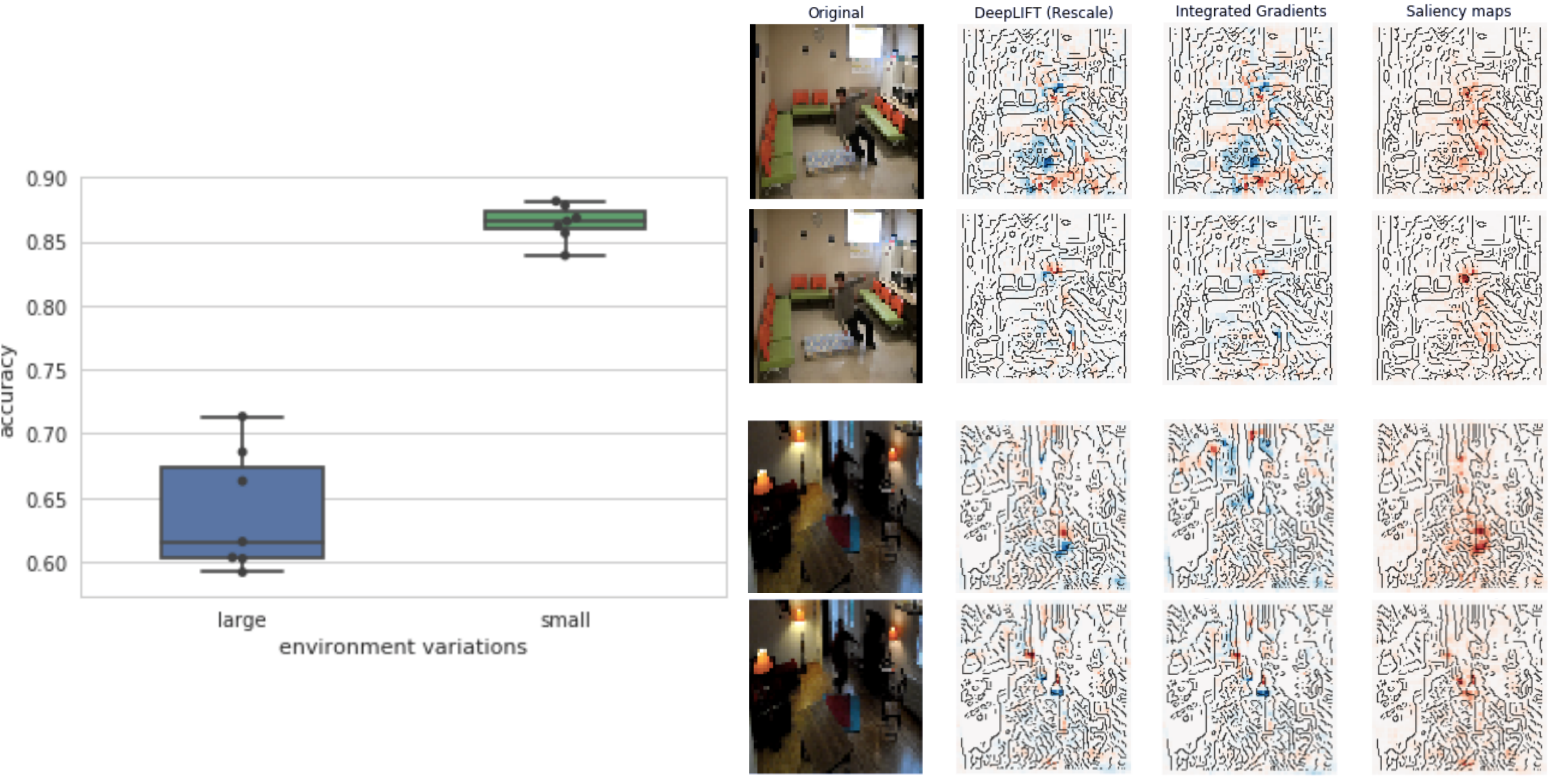}
\caption{From left to right: (1) The quantitative results under large environment variations (the {\em high-level} splits) and small environment variations (the {\em low-level} splits), with the modality {\bf RGB+TimeDifference}. (2) Attribute maps from two testing samples are shown. The first two rows compare the large and small evaluation settings on the same frame in {\em coffee room}, respectively. The last two rows show another comparison on the same frame in {\em home}. }
\label{fig:small_environment_variation}
\end{figure}

In this task, we aim at investigating the influence of training samples recorded from similar environments to the testing samples. Thus, we perform 4-fold cross-validation based on the {\em low-level} dataset splitting, and compare the performances with the 2-fold cross-validation setting (see Task 1). The employed input modality is {\bf RGB+TimeDifference} and the results are shown in Figure \ref{fig:small_environment_variation}. The reason of only using {\bf RGB+TimeDifference} is that other modalities used in Task 1, namely, {\bf TimeDifference} and {\bf Optical Flow}, are environment-independent and cannot reveal the influence of environment variations.

The box plots show that the training recordings from similar environments to testing can largely improve the performances. Indicated by all attribute maps on the right, we can find that the influential pixels noticeably become more human body-centered. 

\paragraph{Discussion} Quantitatively, training recordings similar to the testing recordings are highly favorable. The reason can be revealed from the attribute maps. Specifically, the fact that influential pixels are more concentrated around the human body can also indicate that the fall features are human body-centered. In addition, one can notice that the body-centered influential pixels tend to locate around the contour of the body, instead of directly on the body. This fact may indicate that the body-centered context, or the interaction between the human body and the environment, is a representative feature of fall.

\subsubsection{Task 3: Investigating the human body-centered pattern}

\begin{figure}
\centering
\includegraphics[height=6.2cm]{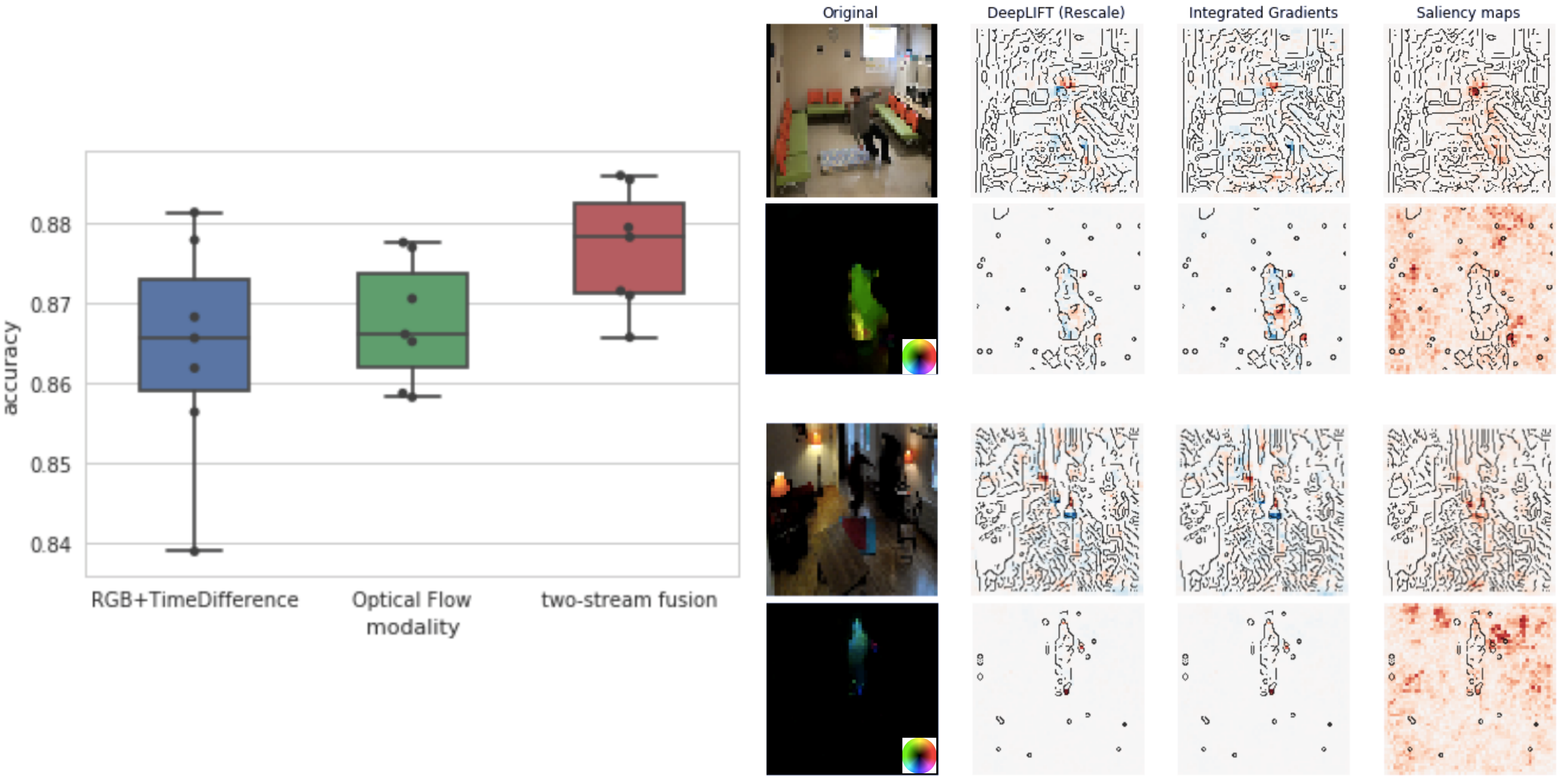}
\caption{From left to right: (1) The quantitative results of different modalities under small environment variations (the {\em low-level} splits). (2) Examples of attribute maps of the two modalities are presented. In particular, the optical flow is visualized using the color coding scheme attached at the bottom-right corner. }
\label{fig:fusion}
\end{figure}

Based on the results in Task 1 and Task 2, we believe that the convolutional net tends to learn body-centered patterns for fall recognition. Here we perform further investigations based on the {\em low-level} data splitting and the {\bf RGB+TimeDifference} and {\bf Optical Flow} modalities, which represent body-centered context and body motion, respectively. Afterwards, we fuse the two modalities following the work of \cite{simonyan2014two}. Specifically, we average the softmax outputs from two streams of CED nets with the same $(k,l)$ values. Figure \ref{fig:fusion} shows the results.

One can see that {\bf Optical Flow} and {\bf RGB+TimeDifference} lead to comparable performances according to the box plots, yet the net with the {\bf Optical Flow} modality behaves more stable than the other case. The fusion results outperform individual modalities. Additionally, from the attribute maps of optical flows we can see that the influential pixels are within the contour of the human body, in contrast to the attribute maps of {\bf RGB+TimeDifference}. One can note that the saliency map is not reliable for {\bf Optical Flow}, since the saliency values are computed as the derivatives of the output w.r.t. the input and zero-value input can cause numerical problems. 

\paragraph{Discussion} A probable reason of the stable performance with {\bf Optical Flow} is that human body motion can represent falls more robustly than the body-centered context, which can be easily influenced by the background information. The superior performance of modality fusion can indicate that body-centered context and body motion are complementary. The complementary property can also be viewed from the attribute maps, since the influential pixels are at different locations. 

\subsubsection{Task 4: Investigating the influence of body pose information}

\begin{figure}
\centering
\includegraphics[height=6.2cm]{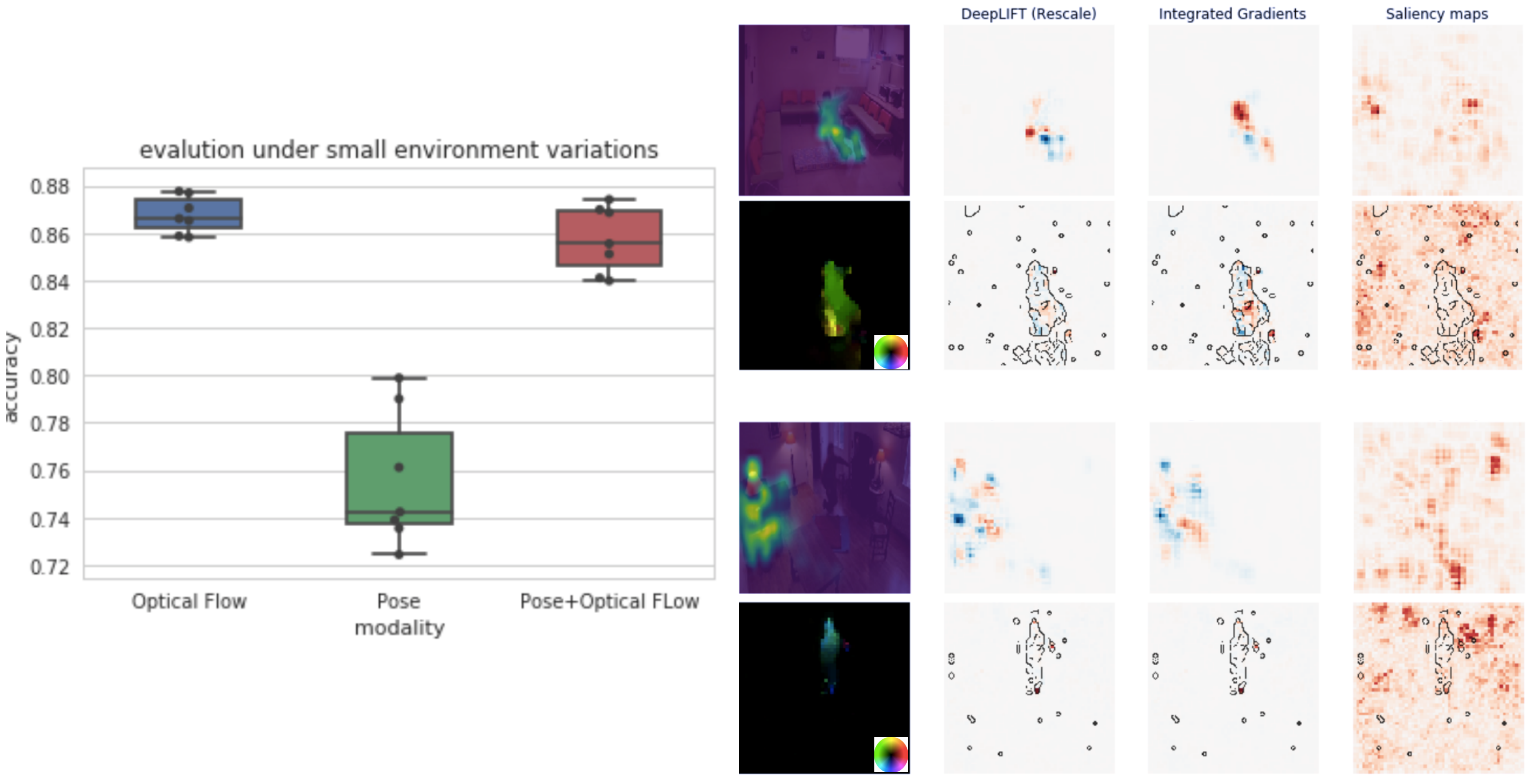}
\caption{From left to right: (1) The quantitative results presented by box plots. (2) The attribute maps of pose and optical flow modalities. The selected frames are the same with previous figures. The pose score map, in which the value increases from blue to yellow, is overlaid with the RGB image only for visualization. The RGB image is not input to the net. }
\label{fig:pose}
\end{figure}

Here we aim at investigating the influence of the 2D pose information. Since motion capture devices are not used in the dataset and no body pose annotations are available, the pose maps are extracted using the pre-trained model associated with \cite{insafutdinov2016eccv} \cite{insafutdinov2017cvpr}. The evaluation is based on the {\em low-level} splitting, as well as the modalities of {\bf Pose}, {\bf Optical Flow} and {\bf Pose+Optical Flow}. The results are shown in Figure \ref{fig:pose}.

One can see that the pose information leads to inferior performances, and also deteriorates the performances of {\bf Optical Flow} when combing flow and pose information. On the right hand, one can find that the influential pixels on the pose score maps mainly locate at the positions the non-zero pose scores. Similar to the optical flow case, the saliency maps of {\bf Pose} are deteriorated by numerical problems. Moreover, from the third row on the right, one can see that the pose estimation is not always reliable.

\paragraph{Discussion}
Pose estimation from images is a challenging problem. Although the state-of-the-art algorithms perform quite well on standard benchmarks, the estimation result is not guaranteed. In case of fall recognition, we can see that incorrect pose estimation can dramatically degrade the performances.

\section{Conclusion and Future work}
\label{sec:conclusion}
In this paper, we aim at investigating the behaviors of the convolutional neural net when conducting fall recognition. To enable frame-wise recognition, we use the convolutional encoder-decoder (CED) architecture and employ a set of net instances. Based on different types of input modalities and dataset splits, our empirical studies show several influential factors of the model performances. In particular, we find that: (1) The net tends to learn body-centered patterns, but cannot eliminate the influence of background information, leading to poor cross-environment generalizability. Therefore, for cross-environment uses in practice, it is better to perform person detection as a pre-processing step, or incorporate a region-of-interest proposing module into an end-to-end model, like the Faster R-CNN model \cite{ren2015faster}.
(2) Training samples captured from the testing environment can considerably improve the performance and encourage the net to encode body-centered context, for which the most influential pixels are located around the body contour. Thus, in practice, we suggest to collect training samples from the deployment environment when possible. 
(3) The human body motion contains representative features of falls robust to environment changes, and influences on fall recognition in a complementary manner with the body-centered context. In this case, we suggest to use the two-stream (the appearance stream and the motion stream) architecture \cite{simonyan2014two} when detecting falls. In addition, since the body-centered context and the body motion are from different image regions, their correlation could be trivial and we probably can effectively fuse the two types of feature vectors only by concatenation or averaging. 
(4) Incorrect pose information can degrade the performances heavily. At the current stage, body pose estimation is a challaging task by itself, and the performances are not guaranted. We hence recommend not to incorporate pose information for fall recognition without additional checking.

Herein we focus on trimmed videos for investigating the net behaviors. Based on the obtained insights, we consider to develop an effective fall detection system based on the CED architecture for untrimmed videos or even streaming data in future.

\paragraph{Acknowledgements}
This work is supported by a grant of the Federal Ministry of Education and Research of Germany (BMBF) for the project of SenseEmotion.

%
%
%
 \bibliographystyle{splncs04}
 \bibliography{reference}
%




\end{document}